\def\year{2022}\relax
\documentclass[letterpaper]{article} 
\usepackage{aaai22}  
\usepackage{times}  
\usepackage{helvet}  
\usepackage{courier}  
\usepackage[hyphens]{url}  
\usepackage{graphicx} 
\usepackage{natbib}  
\usepackage{caption} 
\DeclareCaptionStyle{ruled}{labelfont=normalfont,labelsep=colon,strut=off} 
\usepackage{algorithm}
\usepackage{algorithmic}
\usepackage{amsmath}
\usepackage{amsfonts}
\usepackage{multirow}

\usepackage{newfloat}
\usepackage{listings}
\floatstyle{ruled}
\newfloat{listing}{tb}{lst}{}
\floatname{listing}{Listing}
\title{Unboxing the graph: Interpretable graph neural networks for transport prediction through neural relational inference}
\author{
    Mathias Niemann Tygesen,
    Francisco C. Pereira\equalcontrib,
    Filipe Rodrigues\equalcontrib
}
\affiliations{
    Machine Learning for Smart Mobility, Transport Division, DTU Management\\
    
    Bygningstorvet 116B\\
    2800 Kgs. Lyngby, Denmark\\
    mnity@dtu.dk
}

\begin{document}

\maketitle

\begin{abstract}
Predicting the supply and demand of transport systems is vital for efficient traffic management, control, optimization, and planning. For example, predicting where from/to and when people intend to travel by taxi can support fleet managers to distribute resources; better predicting traffic speeds/congestion allows for pro-active control measures or for users to better choose their paths. Making spatio-temporal predictions is known to be a hard task, but recently Graph Neural Networks (GNNs) have been widely applied on non-euclidean spatial data. However, most GNN models require a predefined graph, and so far, researchers rely on heuristics to generate this graph for the model to use. In this paper, we use Neural Relational Inference to learn the optimal graph for the model. Our approach has several advantages: 1) a Variational Auto Encoder structure allows for the graph to be dynamically determined by the data, potentially changing through time; 2) the encoder structure allows the use of external data in the generation of the graph; 3) it is possible to place Bayesian priors on the generated graphs to encode domain knowledge. We conduct experiments on two datasets, namely the NYC Yellow Taxi and the PEMS road traffic datasets. In both datasets, we outperform benchmarks and show performance comparable to state-of-the-art. Furthermore, we do an in-depth analysis of the learned graphs, providing insights on what kinds of connections GNNs use for spatio-temporal predictions in the transport domain.
\end{abstract}

\section{Introduction}
\label{introduction}
A significant factor in modern intelligent transport systems is preemptively adapting to changes. This could be changes in taxi demand in a city or speeds on a city's road network. By matching the actual taxi demand, planners can best utilize the fleet of taxis, both lowering costs and waiting times for users. However, doing so requires a good demand model that allows transport planners to best plan the supply and adopt early interventions when the demand suddenly changes. However, ride-hauling transport systems' demand prediction has highly non-euclidean spatial dependencies between zones. This could be dependencies between residential and industrial zones, between zones with many hotels and zones with many tourist attractions, modal interchange stations and other areas, etc. Similarly, for road networks, vehicles can be rerouted around those areas to optimize traffic flow better by predicting traffic jams. 

In recent years, Graph Neural Networks (GNNs) have shown great performance on different tasks in non-euclidean spaces \cite{BronsteinGeometricDeeplearning}. Also, in transport systems, GNNs have shown remarkable performance in both link-based flow/speed prediction and demand prediction for different transport systems  \cite{Li2018DiffusionCR}, \cite{Ye2021CoupledLG}. While the road network is the inherent graph for many flow/speed prediction tasks, the problem is that this graph may not be optimal as a road link could have higher-order dependencies with road links further away. Furthermore,  creating the true graph for a road network might require error-prone map-matching procedures. For ride hauling systems, there is no inherent graph, as the system is divided into zones or stations with no natural connections. To address this, researchers have previously used different heuristic methods to create a graph of the system for the GNNs to use. While these heuristic graphs allow to include domain knowledge, they are a modeling choice and might not depict the true correlation or dependencies between roads, zones, or stations. On top of the spatial dependencies, transport systems also have strong temporal dependencies with daily and weekly trends in demand. Previously, this has been modeled by including Recurrent Neural Networks to take these trends into account on a per-node basis. However, heuristically-created graphs are static through time which means that models have the same assumption regarding correlations between zones for all time steps. Unfortunately, the true dependencies in a transport system might change over time. For example, it seems highly plausible that the dependencies between a residential zone and an industrial zone with many jobs show clear dependencies to the industrial zone in the morning and back to the residential zone in the afternoon, but none during weekends or nights. \\
In this paper, we utilize Neural Relational Inference \cite{kipf_nri} (NRI), which uses a Variational Auto-Encoder (VAE) structure to learn a distribution over the possible edges in the graph. This way, the model infers the graph that leads to the best predictions. Under the assumption that the true dependency graph will lead to the most accurate predictions, this inferred graph will approximate the true dependencies. Another benefit of this approach is that it enables the researcher to put a Bayesian prior over the latent graph distribution, enabling them to apply domain knowledge without limiting the model's flexibility. Furthermore, unlike others, our model is not constrained to a specific number of nodes. Hence, the model can be used in settings where the number of nodes is different throughout the dataset, i.e., a road network with a varying number of loop detectors over a few years. Therefore, the model could also be used in a transfer learning setup where the model is trained on a transport system with a large amount of data and later fine-tuned to a scenario with sparse data, even if the number of nodes is different. \\
On top of conducting experiments that show the model's predictive capabilities, we also analyze the inferred graphs, showing the types of connections the model prefers in different transport systems. To the best of the authors' knowledge this is the first in depth analysis of inferred graphs in transport data. We do this to show that the learnable adjacency matrix is a step towards more interpretable models and to show what kind of connections models learn to optimize predictions. \\
In summary, our contributions are:
\begin{itemize}
\item a proposal for automatic inference of a dynamic graph adjacency matrix for use in a GNN for transport data, that shows performance competitive (or, in some cases, superior) with other state-of-the-art GNN approaches;
\item a study demonstrating significant performance improvements in comparison with the mostly used heuristics in GNN graph structure design;
\item an application of the method both from a transport supply and demand perspectives;
\item a thorough analyses on how the training of the model affects the learned matrices and how the learned matrices shows what the model focuses on for better predictions.

 %
\end{itemize}

\section{Literature review}
In this section we review deep learning transport research and particularly focus on the graph inference problem in using GNNs on transport data.

\subsection{Deep learning in transport research}
Prediction tasks in transport systems have seen much research, and many different methods have been used on the problem. Historically, classical statistical methods like ARIMA \cite{Williams2003} and classical machine learning methods like Gaussian Processes \cite{Diao2019} have been used, however these face important challenges when considering large scale urban network spatial and temporal dependencies. Variants exist, like Vector ARIMA (VARIMA \cite{barcelo2011compositional}) or Multi-output GPs \cite{rodriguez2017urban} that aim to cover such dependencies, but they are either too rigid (once calibrated, parameters are fixed) for real-world cities, or computationally challenging \cite{liu2020gaussian}. 
However, recent developments in  deep learning paved the way for efficient and flexible modeling of spatial and temporal dependencies in transport data. For the temporal dependencies, Recurrent Neural Networks (RNN) have been utilized \cite{Xu2018RealTimePO}. Initially, the spatial dependencies have been modeled using Convolutional Neural Networks (CNN) \cite{Zhang2016DNNbasedPM}, yet CNN's require the data to follow a grid data structure which not all transport data lends itself to. Therefore, researchers have recently focused on utilizing graph neural networks (GNN) in transport contexts \cite{Li2018DiffusionCR}. GNNs' ability to capture both short- and long-distance spatial relationships makes it especially attractive for transport applications which is evident by the large amount of research published. See \citet{jiang2021graph} and \citet{Yin2021} for more exhaustive surveys.

\subsection{Graph inference in transport research}
Most GNN models require a predefined graph. For problems with an intrinsic graph structure like road networks, molecules, or social networks, there exists a natural choice of graph for the GNN to use. However, this graph may not be optimal for accurate predictions. For problems with no intrinsic structure like station or zone-based ride hauling demand prediction, researchers have no obvious graph for the GNN and have to resort to heuristics to create a graph. Examples of such heuristics include graphs based on geospatial distance, neighboring zones, time-series correlation, or zone functional similarity (e.g., based on points-of-interest, land-use). Even after the choice of metric to use, the researcher also has a choice of how to translate the similarities between nodes into a graph, i.e., choose a cut-off similarity to determine what is an edge and what is not. While this allows applying domain knowledge in the modeling process, these graphs are fixed and might not reflect the true underlying dependencies in the data, leading to bias in the model and lower predictive performance. Multiple approaches have been applied in transport research to try and amend this problem. In \cite{Bai2020AdaptiveGC} and \cite{Wu2019GraphWF}, the authors randomly initialize one or two matrices of learnable node embeddings and multiply them together, followed by an activation function and softmax to get an adjacency matrix. This adjacency matrix can then be learned during training using stochastic gradient descent. In \cite{Tang2020AGT}, the authors calculate attention weights between all nodes in the network and use those as weights in the adjacency matrix. \citet{Ye2021CoupledLG} start with a similarity-based adjacency matrix but use a stack of learnable mapping functions to generate multiple adjacency matrices. The authors then aggregate the embeddings from all the graphs for the final output.
Our method varies from previous methods used in transport research in multiple ways. First, we utilize the VAE structure of the NRI model. This means that we have a fully learnable adjacency graph that is not locked to the number of nodes in the graph. Hence the model can be used in a transfer- or meta-learning setting. Another benefit of the VAE structure is that we can put priors on the learned graph. This can be a way for experts to apply their domain knowledge without limiting the flexibility of the models. We also utilize the GN structure \cite{Battaglia_relational_biases} which allows for both the encoder and the decoder to utilize global features.

\section{Definitions}

\subsection{Graph definition}
A graph is defined as a 3-tuple $G = (\boldsymbol{u}, V, E)$. $V = \{\boldsymbol{v}_i\}_{i=1:N^v}$ is the set of $N^v$ vertices and $\boldsymbol{v}_i$ are the $i$'th node features. $E=\{(\boldsymbol{e}_k, r_k, s_k)\}_{k=1:N^e}$ is the set of $N^e$ edges were $\boldsymbol{e}_k$ are the edge features and $r_k$ is the receiver nodes index and $s_k$ is the sender nodes index. Note that for brevity we will use $\boldsymbol{e}_{i,j}$ to refer to $\boldsymbol{e}_k$ where $s_k = i$ and $r_k = j$. $\boldsymbol{u}$ is a global feature for the entire graph, e.g. air temperature for a traffic system. Often the connections of the graph are represented in an adjacency matrix $A \in \mathbb{R}^{n \times n}$ where $A_{ij} = 1$ if $\boldsymbol{e}_{(i,j)} \in E$ and 0 otherwise. Note here that the adjacency matrix notation does not contain the edge features but only the connections.

\subsection{Graph Neural Networks}
Graph neural networks have seen much research in recent years. Therefore there are many types of GNNs \cite{KipfGCN, HamiltonYL17Sage, GilmerSRVD17MPNN}. The Graph Network framework \cite{Battaglia_relational_biases} unifies most of these networks. In the GN framework, a GNN consists of GN blocks, each consisting of 3 update functions and three aggregation functions
\begin{equation}
\label{eq:GN_formulation}
\begin{aligned}
&\mathbf{e}_{k}^{\prime}=\phi^{e}\left(\mathbf{e}_{k}, \mathbf{v}_{r_{k}}, \mathbf{v}_{s_{k}}, \mathbf{u}\right) & \overline{\mathbf{e}}_{i}^{\prime}=\rho^{e \rightarrow v}\left(E_{i}^{\prime}\right) \\
&\mathbf{v}_{i}^{\prime}=\phi^
{v}\left(\overline{\mathbf{e}}_{i}^{\prime}, \mathbf{v}_{i}, \mathbf{u}\right) & \overline{\mathbf{e}}^{\prime}=\rho^{e \rightarrow u}\left(E^{\prime}\right) \\
&\mathbf{u}^{\prime}=\phi^{u}\left(\overline{\mathbf{e}}^{\prime}, \overline{\mathbf{v}}^{\prime}, \mathbf{u}\right) & \overline{\mathbf{v}}^{\prime}=\rho^{v \rightarrow u}\left(V^{\prime}\right),
\end{aligned}
\end{equation}
where $E_{i}^{\prime}=\left\{\left(\mathbf{e}_{k}^{\prime}, r_{k}, s_{k}\right)\right\}_{r_{k}=i,\ k=1: N^{e}}$, $V^{\prime}=\left\{\mathbf{v}_{i}^{\prime}\right\}_{i=1: N^{v}}$ and  $E^{\prime}=\bigcup_{i} E_{i}^{\prime}=\left\{\left(\mathbf{e}_{k}^{\prime}, r_{k}, s_{k}\right)\right\}_{k=1: N^{e}}$. The edge and node update functions $\phi^e$ and $\phi^v$ are mapped across all edges and nodes respectively and the global update $\phi^u$ function is applied once. $\rho$ is a permutation invariant aggregation function that can take varying number inputs, e.g. elementwise summation, mean or maximum.

\subsection{Spatio-temporal prediction in transport systems}
The task of spatio-temporal prediction in transport systems is to predict the system's future states, given the historical states. In order for GNN's to be used for the prediction, the system needs to be in a form with nodes. For systems with clear nodes, i.e., stations in a metro network or loop sensors along a highway, this step is trivial. However, for problems with no clear nodes, like a ride-hauling service, the data has to be preprocessed into a node-based form first. This can be done by using predefined zones and cluster the observations into these zones. Another method is using clustering algorithms to cluster the observations into ``virtual" zones \cite{Ye2021CoupledLG}. Here we will only use data either in a node form or clustered into predefined zones by the data provider. \\
\\
For a transport system with $N$ nodes let $\boldsymbol{x}^{t} \in \mathbb{R}^{N \times c}$ denote the feature matrix at time step $t$. Here $c$ denotes the feature dimension. Then the task of spatio-temporal prediction is given the last $P$ timesteps, predict the next $Q$ time steps can be denoted as
\begin{equation}
    \boldsymbol{x}^{t - P + 1: t} \xrightarrow{\mathcal{F}} \boldsymbol{x}^{t +1: t + Q},
\end{equation}
where $\mathcal{F}$ is a mapping function, $\boldsymbol{x}^{t - P + 1: t} \in \mathbb{R}^{P \times N \times c}$ and $\boldsymbol{x}^{t +1: t + Q} \in \mathbb{R}^{Q \times N \times c}$. \\
Given that $\mathcal{F}$ is a GN following the structure from Equation \ref{eq:GN_formulation}, taking in general graph features, the formulation becomes
\begin{equation}
    [\boldsymbol{x}^{t - P + 1: t}, \boldsymbol{u}^{t - P + 1 : t}, G] \xrightarrow{\mathcal{F}_{GN}} \boldsymbol{x}^{t +1: t + Q},
\end{equation}
where $G$ is the graph structure and $\boldsymbol{u}^{t} \in \mathbb{R}^{c_u}$ is the general graph features, with $c_u$ dimensions, of time step $t$, hence $\boldsymbol{u}^{t - P + 1 : t} \in \mathbb{R}^{P \times c_u}$.

\section{Recurrent Spatio-temporal Graph Neural Network with Neural Relational Inference}
\label{sec:model}
\begin{figure*}[t]
    \centering
    \includegraphics[width=0.95\textwidth]{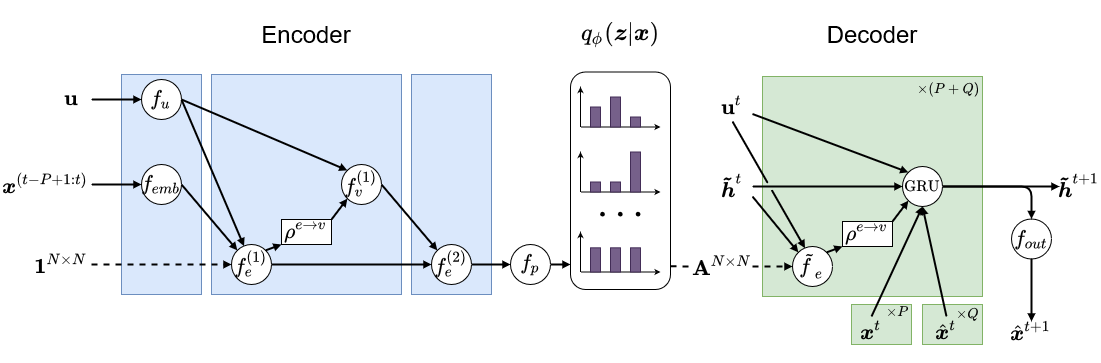} 
    \caption{The full model structure. First, the encoder takes in the global feature as well as the historical data and outputs parameters for the latent distribution $q_\phi(\boldsymbol{z}| \boldsymbol{x})$. Then the concrete distribution is used to get a sample from the latent space. This sampled graph is then used in the recurrent decoder. The decoder takes in the true data $\boldsymbol{x}^t$ for the first $P$ steps and then the last prediction $\hat{\boldsymbol{x}}^t$ for the last $Q$ steps}
    \label{fig:full_model}
\end{figure*}
In this section, we present the Neural Relational Inference GNN model. The model follows the basic structure from \cite{kipf_nri} but inspired by \cite{Battaglia_relational_biases} a global graph feature is added. This global value is then used to condition the edge-to-node and node-to-edge transformations in both the encoder GNN and the decoder GNN. Furthermore, we fully utilize the prior by constructing a structured prior using domain knowledge. The model can be seen as a VAE \cite{Kingma2014}. With our extensions, the model maximizes the ELBO:
\begin{equation}
\begin{aligned}
    \mathcal{L} = \mathbb{E}_{q_\phi(\boldsymbol{z}| \boldsymbol{x}^{t-P+1:t}, \boldsymbol{u})}&[\log p_\theta(\boldsymbol{x} | \boldsymbol{z}, \boldsymbol{u})] \\
    &- \text{KL}[q_\phi(\boldsymbol{z}| \boldsymbol{x}^{t-P+1:t}, \boldsymbol{u}) || p(\boldsymbol{z})],
\end{aligned}
\end{equation}
where
\begin{equation}
    p_\theta(\boldsymbol{x} | \boldsymbol{z}, \boldsymbol{u}) = \prod_{t'=t+1}^{t+Q} p_{\theta}(\boldsymbol{x}^{t'+1} | \boldsymbol{x}^{t'}, \ldots, \boldsymbol{x}^{t+1}, \boldsymbol{z}, \boldsymbol{u})
\end{equation}
is the decoder that predicts the next step given the previous observations, the latent graph structure, and the global feature,
and 
\begin{equation}
   p(\boldsymbol{z}) = \prod_{i\neq j} p(\boldsymbol{z}_{(i,j)})
\end{equation}
is a factorized prior on the latent graph. $p(\boldsymbol{z}_{(i,j)})$ can either be the same prior for all the edges or be structured with a different distribution for each edge based on prior knowledge. \\

\noindent
\textbf{Encoder} \\
\noindent
The encoder $q_\phi(\boldsymbol{z}| \boldsymbol{x})$ takes as input the historical data $\boldsymbol{x}^{t-P+1:t}$ and global feature $\boldsymbol{u}$. The encoder follows the structure of a GNN running on a fully connected graph $\textbf{1}^{N \times N}$. In the GN framework, the encoder can be seen as three GN blocks stacked after each other: an independent recurrent block that updates the embedding of the nodes and global feature, a block that creates edge features and updates node features, and finally a block that updates the edge features. After that, a fully connected neural network takes the final edge embeddings and outputs the probabilities for the categorical distribution in the latent space. Let $\boldsymbol{x}_j$ denote the features of the $j$th node, then the node and edgewise computations can be written as
\begin{equation}
    \begin{aligned}
    \boldsymbol{h}_j^{(1)} &= f_{emb}(\boldsymbol{x}^{t-P+1:t}_j) \\
    \boldsymbol{u}^{(1)} &= f_{u}(\boldsymbol{u}) \\
    v \rightarrow e: \boldsymbol{h}_{(i,j)}^{(1)} &= f_e^{(1)}(\boldsymbol{h}_i^{(1)}, \boldsymbol{h}_j^{(1)}, \boldsymbol{u}^{(1)}]) \\
    e \rightarrow v: \boldsymbol{h}_j^{(2)} &= f_v^{(1)}([\sum_{i \neq j} \boldsymbol{h}_{(i,j)}, \boldsymbol{u}^{(1)}]) \\
    v \rightarrow e: \boldsymbol{h}_{(i,j)}^{(2)} &= f_e^{(2)}([\boldsymbol{h}_i^{(2)}, \boldsymbol{h}_j^{(2)}, \boldsymbol{h}_{(i,j)}^{(1)}]) \\
    \boldsymbol{h}_{(i,j)}^{(3)} &= f_p(\boldsymbol{h}_{(i,j)}^{(2)}),
    \end{aligned}
\end{equation}
where the $f$'s are fully connected neural networks.  The structure of the encoder is depicted in Figure \ref{fig:full_model}. \\

\noindent
\textbf{Latent space} 
\label{sec:latent_space}\\
\noindent
In order to be able to backpropagate through the latent space $q_\phi(\boldsymbol{z}_{ij}|\boldsymbol{x})$, we do a continuous relaxation using the Gumbel distribution \cite{Jang2017, Maddison2017}. This way, the samples are drawn as:
\begin{equation}
    \boldsymbol{z}_{i,j} = \text{softmax}((\boldsymbol{h}_{(i,j)}^{(3)} + \boldsymbol{g}) / \tau),
\end{equation}
where $g$ is a vector of $k$ i.i.d samples from a Gumbel(0,1) distribution and $\tau$ is a temperature that controls the "smooth" relaxation of the samples. We collect these samples into an adjacency matrix $\textbf{A}^{N  \times N}$ s.t. $\textbf{A}^{N  \times N}_{i,j} = \textbf{z}_{i,j}$. \\

\noindent
\textbf{Decoder} \\
\noindent
In order for the decoder to be able to model the temporal dependencies in $p_{\theta}(\boldsymbol{x}^{t'+1} | \boldsymbol{x}^{t'}, \ldots, \boldsymbol{x}^{t+1}, \boldsymbol{z}, \boldsymbol{u})$ we utilize a GRU cell in the decoder. The decoder can be seen as a recurrent GN block where the hidden state in the GRU cell is the node embeddings. The block first does an edge embedding using the hidden state and the graph $\textbf{A}^{N  \times N}$ sampled from the latent space. Then, the edge embeddings are used to do a node embedding update using the GRU cell. The GRU also takes the observed node features and global feature as input and outputs a hidden state for the next time step. This hidden state is then both sent to the next iteration and used to create predictions for the next time step.
\begin{equation}
    \begin{aligned}
    v \rightarrow e: \Tilde{\boldsymbol{h}}^t_{(i,j)} &=\sum_k z_{ij,k} \Tilde{f}_e^k([\Tilde{\boldsymbol{h}}_i^t, \Tilde{\boldsymbol{h}}_j^t, \boldsymbol{u}^t]) \\
     e \rightarrow v: \text{MSG}_j^t &= \sum_{(i\neq j)} \Tilde{\boldsymbol{h}}_{(i,j)}^t \\
    \Tilde{\boldsymbol{h}}_j^{t+1} &= \text{GRU}([\text{MSG}_j^t, \boldsymbol{x}_j^t, \boldsymbol{u}^t], \Tilde{\boldsymbol{h}}_j^t) \\
    \boldsymbol{\mu}^{t+1} &= \boldsymbol{x}_j^t + f_{out}(\Tilde{\boldsymbol{h}}_j^{t+1}) \\
    p(\boldsymbol{x}^{t+1} | \boldsymbol{x}^t, \boldsymbol{z}) &= \mathcal{N}(\boldsymbol{\mu}^{t+1}, \sigma^2I). 
    \end{aligned}
\end{equation}
The tilde is used to denote the decoders hidden states and edge embedding networks such that they can be distinguished from the encoders.

\subsection{Training}   
During training, we first send the historical data $\boldsymbol{x}^{t-P+1:t}$ and the global features $\boldsymbol{u}$ into the encoder. Note that we provide the entire global feature as input to the model. We do this as we will use global features with external predictions that we assume to be of high enough quality to work with the model. However, it is easy to constrain the encoder to only historical global features, i.e., $\boldsymbol{u}^{t-P+1:t}$. The encoder then outputs the latent distribution over graph edges $q_\phi(\boldsymbol{z}|\boldsymbol{x}, \boldsymbol{u})$, from which we sample a graph. We then use the sampled graph for the decoder. The decoder runs $P+Q$ iterations. For the first $P$ iterations, we input the true observations $\boldsymbol{x}^t$ into the decoder, and for the last $Q$ iterations, we input the predictions $\boldsymbol{\hat{x}}^t$ into the decoder. \\
The loss then consists of a KL divergence part and a reconstruction part. The KL divergence part is given by
\begin{equation}
\begin{aligned}
    \text{KL}[q_\phi(\boldsymbol{z}&| \boldsymbol{x}, \boldsymbol{u}) \| p(\boldsymbol{z})] = \\
    &\sum_z q_\phi (\boldsymbol{z} | \boldsymbol{x}, \boldsymbol{u}) \log \left( \frac{q_\phi (\boldsymbol{z} | \boldsymbol{x}, \boldsymbol{u})}{p(\boldsymbol{z})} \right),
\end{aligned}
\end{equation}
and the reconstruction part is given by negative log-likelihood (NLL) summed over nodes and prediction time steps:
\begin{equation}
    \log p_\theta(\boldsymbol{x} | \boldsymbol{z}, \boldsymbol{u}) = - \sum_j^{N} \sum_{t=t}^{t+Q} \left(\frac{\| \boldsymbol{x}_j^t - \boldsymbol{\hat{x}}_j^t\|^2  }{2\sigma^2} + \frac{1}{2} \log(2 \pi \sigma)\right).
    \label{eq:loss}
\end{equation}
The value of $\sigma$ can be seen as a hyperparameter determining how much the negative log-likelihood should penalize errors. This can be used to scale the ratio between the negative log-likelihood and the KL divergence, thereby putting more or less emphasis on the prior.

\section{Experiments}
In this section, we perform experiments on two different datasets. The source code for all experiments is available.\footnote{https://github.com/MathiasNT/NRI\_for\_Transport}
\subsection{Datasets}
\label{sec:datasets}
\begin{figure}[t]
    \centering
    \includegraphics[width=1\columnwidth]{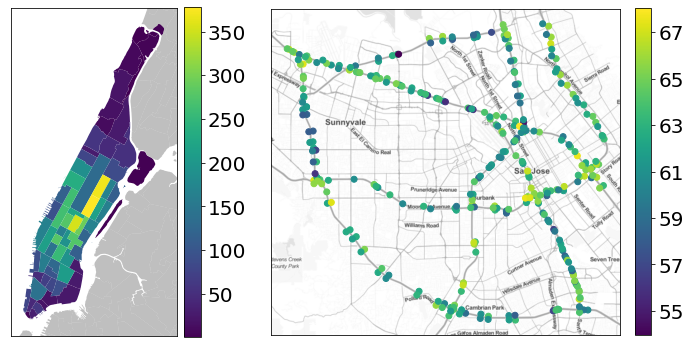} 
    \caption{Left shows the 66 taxi zones of Manhattan color-coded based on the mean hourly activity over the data period. Right shows the location of the 325 loop sensors in the PEMS data. The color is the mean speed at the sensor over the data period. The location of the sensors has been altered slightly to show sensors placed on opposite sides of the same freeway.}
    \label{fig:data_plot}
\end{figure}

The experiments are done on two open-source datasets.
\begin{itemize}
    \item \textbf{NYC Yellow Taxi Data\footnote{https://www1.nyc.gov/site/tlc/about/tlc-trip-record-data.page}}: The dataset contains logs for all of the yellow taxis driving around New York. The full year of 2019 is used, and we extract the data from Manhattan. The data contains 84 million raw trips. Each trip logs the starting zone, ending zone, time of the start and end, trip distance, trip duration, and trip cost.\\
    \item \textbf{PEMS\footnote{https://github.com/liyaguang/DCRNN}}: The dataset contains logs of traffic speeds from 325 loop sensors around the Bay Area collected by the California Transportation Agencies (Cal-Trans) from January 1st, 2017 to May 31th 2017.
\end{itemize}

\subsection{Data Preprocessing}
The NYC Yellow Taxi dataset is first cleaned for erroneous logs. We remove all trips with:
\begin{itemize}
    \setlength\itemsep{0em}
    \item A distance shorter than 0.1 mile;
    \item A duration shorter than 1 minute;
    \item Free trips or trips with a negative cost;
    \item Trips with errors in the timestamps.
\end{itemize}
After cleaning, the dataset contains 71 million trips. We then aggregate the data into pickups and dropoffs in each NYC taxi zone per hour. We create splits of 60 hours with 48 hours for burn-in and 12 hours for prediction, i.e., given the last two days, the task is to predict the pickups and dropoffs in each zone for the next 12 hours. We then use a 0.8/0.1/0.1 train, validation, and test split and normalize based on the data in the train set.

The PEMS dataset is taken from \cite{Li2018DiffusionCR} and hence follows the same train, validation and test splits and is split into vectors of 24 hours with 12 hours for burn-in and 12 hours for prediction. We also standardize the data before training.

\subsection{Experiments on the NYC Yellow Taxicab dataset}
In this section, we do experiments on the NYC Yellow Taxicab dataset. First, we compare the models using different kinds of fixed adjacency matrices with full models with learnable, dynamic adjacency matrices. The full models are trained using different priors. Then we analyze the learned adjacency matrices, using the best model, to see what kind of relations the learned adjacency matrices encode.\\
\\
\noindent
\textbf{Comparing heuristics and learned adjacency matrices} \\
\noindent

\begin{table}[t]
\begin{tabular}{l|llll}
Model & MAE & RMSE & MAPE & PCC \\ \hline
Fixed empty & 23.31 & 49.37 & 75.0\% & 0.943 \\
Fixed full & 19.41 & 32.47 & 43.4\% & 0.976 \\
Fixed local & 22.80 & 37.91 & 50.0\% & 0.967 \\
Fixed DTW & 23.21 & 38.65 & 49.3\% & 0.966 \\
NRI local & 18.36 & 31.69 & 36.8\% & 0.978 \\
NRI unif & 17.54 & 31.69 & 36.8\% & 0.977 \\
NRI DTW & 17.65 & 30.44 & 33.2\% & 0.978
\end{tabular}
\caption{The table shows the performance on the NYC Yellow Taxicab data of the models with different kinds of fixed adjacency matrices and models with learned dynamic adjacency matrices with different priors.}
\label{tab:taxi}
\end{table}

\begin{figure}[]
    \centering
    \includegraphics[width=1\columnwidth]{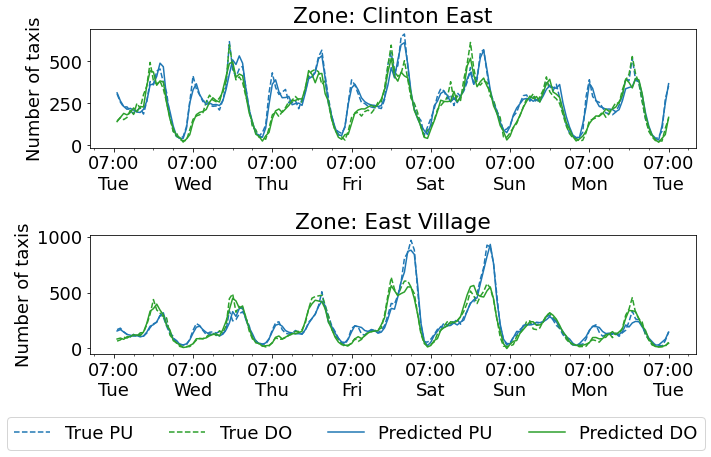} 
    \caption{The figure shows the predictions of dropoffs (DO) and pickups (PU) along with the true pickups and dropoffs for a full week of the test set.}
    \label{fig:taxi_preds}
\end{figure}

We run experiments on the New York City Yellow Taxi Cab dataset, comparing the inferred adjacency matrix against commonly used adjacency matrices based on heuristics. The baseline heuristic adjacency matrices are:
\begin{itemize}
    \item \textbf{Local}. Each zone is connected only to its neighboring zones. Two zones are defined as neighbors if they share a border.
    \item \textbf{DTW}. Each zone is connected only to the zones that have the most similar time series. The similarity between each zone is computed using Dynamic Time Warping (DTW) on its time series in the training dataset. The adjacency matrix is then quantized by the 90th percentile to get a binary adjacency matrix.
    \item \textbf{Empty}. An empty adjacency matrix, meaning that the zones do not share information. The model then becomes an RNN looking at each zone separately.
    \item \textbf{Full}. A full adjacency matrix, meaning that all zones are connected to all other zones.
\end{itemize}

For the models with a fixed adjacency matrix, we use only the decoder part of the model as seen in Figure \ref{fig:full_model}. Then instead of sampling a new graph from $q_\phi(\boldsymbol{z}|\boldsymbol{x})$, we input the fixed adjacency matrix for all steps. For the full NRI models, we use the full model with three different priors: the local and DTW adjacency matrices as explained above and a uniform prior that puts a prior of 90\% probability of no edge on all possible edges. For the NYC Yellow Taxicab dataset, we found a random initialization of the encoder to be too noisy and led to bad training of the model. Therefore, we pre-train the encoder for 30 epochs using only the KL divergence part of the loss. This means that when the model starts full training, the initial outputs of the encoder are similar to the prior.

First, we compare the model with a fixed encoder against the model with the VAE structure with different kinds of priors on the learned adjacency matrix. For the fixed model, the different kinds of adjacency matrices are the local adjacency and the DTW adjacency matrices explained above, along with a full and empty adjacency matrix. The results can be seen in Table \ref{tab:taxi}. As can be seen in the table, among the fixed models, the full adjacency matrix achieves the best performance while the empty adjacency matrix achieves the worst performance. This shows that the model benefits from sending information among the nodes. For the heuristic-based graphs, the local graph achieves the best performance, while the DTW based graph significantly outperforms the empty graph on all metrics except for MAE. When we look at the full NRI models, we can see that all models outperform the fixed encoder approaches, including those with a full adjacency matrix. This suggests that the model is able to learn which edges are important and send more specific messages along these edges. We can also see that the model with a uniform prior achieves the best MAE score while the model with the DTW prior achieves the best performance in all of the other metrics. All of the models have been trained with the $\sigma$ value in the loss (Equation \ref{eq:loss}), that allows the model to overrule the prior. As such, the prior works more as a guide for the training of the model. This is especially true when using the pretraining for the encoder, as the initial adjacency matrix will then match the prior as well.

Next, we look at the predictions made by the best model, the NRI DTW. As can be seen in Figure \ref{fig:taxi_preds}, the model can capture the daily trends. This is especially clear in the Clinton East zone, where each weekday clearly consists of two rush hour peaks. Furthermore, the model is also able to capture weekly trends, as can be seen in the East Village zone, where there is a huge pickup peak during weekend evenings. \\

\noindent
\textbf{Analyzing the learned adjacency matrix} \\
\noindent
\begin{figure}[]
    \centering
    \includegraphics[width=1\columnwidth]{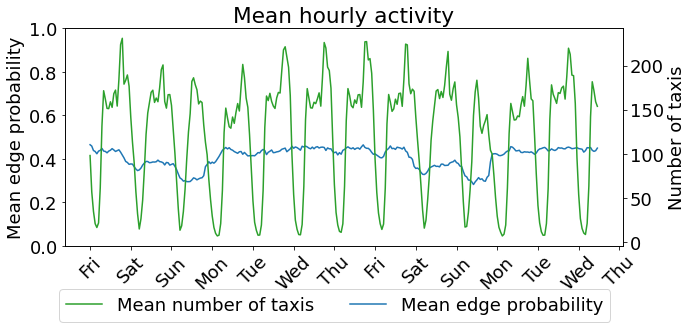} 
    \caption{The mean number of taxis actions (both PUs and DOs) over all zones along with the mean edge probability of the adjacency matrix.}
    \label{fig:taxi_mean_probs}
\end{figure}

\begin{figure}[]
    \centering
    \includegraphics[width=1\columnwidth]{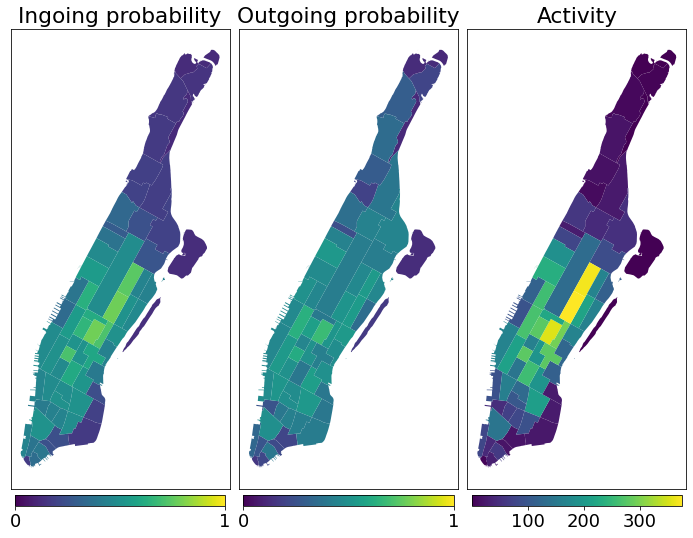} 
    \caption{Mean over time for ingoing edge probability, outgoing edge probability and activity for each zone on Manhattan.}
    \label{fig:taxi_mean_in_out}
\end{figure}

Now, we look at the best performing NRI model, the NRI DTW, to see what kinds of relationships are put in the adjacency matrix. In Figure \ref{fig:taxi_mean_probs}, we show the time series of the mean taxi actions per hour of the entire Manhattan, i.e., the mean of pickups and dropoffs for all zones, along with the mean edge probability for the learned adjacency matrix. As can be seen in the figure, the mean edge probability is constant during the weekdays but has apparent dips during the weekend. This indicates that the messages are optimized for predicting weekday trends, and the model thus sends fewer messages during weekends. Looking at Figure \ref{fig:taxi_mean_probs}, we can also see why this makes sense for the model. First off, there are more weekdays than days on the weekend; secondly, we can see in the figure that, on average, there are fewer taxi trips on weekends than on weekdays. As the model is trained using the loss in Equation \ref{eq:loss}, which puts higher emphasis on large observations, it makes sense for the model to focus on weekdays instead of weekends.

We also analyze which zones send and receive messages. In Figure \ref{fig:taxi_mean_in_out} we show three maps of Manhattan. In the leftmost map, each zone is color-graded according to the mean ingoing edge probability for that zone, in the middle map, we show the mean outgoing edge probabilities, and in the rightmost, we show the mean activity of the zone. As can be seen in the figure, the mean ingoing edge probability is higher for zones with higher mean activity, while the mean outgoing edge probability is more constant across Manhattan, except for outlier zones. This shows that the model learns to send more messages to zones with high activity. This is, again, a result of the emphasis the loss in Equation \ref{eq:loss} puts on large observations. Hence the model learns to create messages that help the most for these zones and sends them to them.

\subsection{Experiments on PEMS dataset}
In this section, we do experiments on the PEMS dataset. We again compare the full model against baselines using fixed heuristic-based adjacency matrices and another graph-based model from the literature. Then, we analyze the learned adjacency matrices to see what kind of relation the model finds important.

\begin{table*}[]
\centering
\begin{tabular}{l|l|llllll}
Time horizon / Model & Metric & DCRNN & Lag & Fixed empty & Fixed full & Fixed local & NRI local prior \\ \hline
\multirow{3}{*}{15 min} & MAE & 1.38 & 1.31 & 1.05 & 1.02 & 1.04 & 1.02 \\
 & RMSE & 2.95 & 2.80 & 2.05 & 1.99 & 2.02 & 2.00 \\
 & MAPE & 2.9\% & 2.6\% & 2.1\% & 2.0\% & 2.1\% & 2.0\% \\ \hline
\multirow{3}{*}{30 min} & MAE & 1.74 & 1.65 & 1.39 & 1.34 & 1.36 & 1.35 \\
 & RMSE & 3.97 & 3.79 & 3.01 & 2.84 & 2.90 & 2.84 \\
 & MAPE & 3.9\% & 3.4\% & 3.0\% & 2.8\% & 2.9\% & 2.8\% \\ \hline
\multirow{3}{*}{60 min} & MAE & 2.07 & 2.18 & 1.93 & 1.8 & 1.84 & 1.78 \\
 & RMSE & 4.74 & 5.19 & 4.31 & 3.90 & 4.07 & 3.89 \\
 & MAPE & 4.9\% & 4.7\% & 4.5\% & 4.0\% & 4.3\% & 4.0\%
\end{tabular}
\caption{Results of different models on the PEMS dataset. Results for the DCRNN model is taken from \cite{Li2018DiffusionCR}.}
\label{tab:pems}
\end{table*}

\begin{figure}[]
    \centering
    \includegraphics[width=0.7\columnwidth]{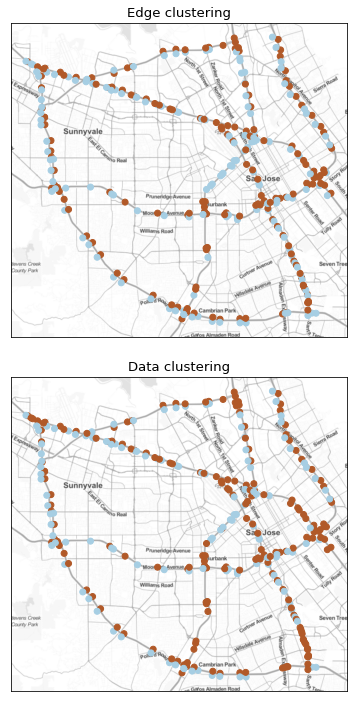} 
    \caption{Top: Clustering based on the time series of learned connections for each sensor. Bottom: Clustering based on the time series of traffic speeds at each sensor.}
    \label{fig:pems_overall_cluster}
\end{figure}

\begin{figure}[]
    \centering
    \includegraphics[width=0.8\columnwidth]{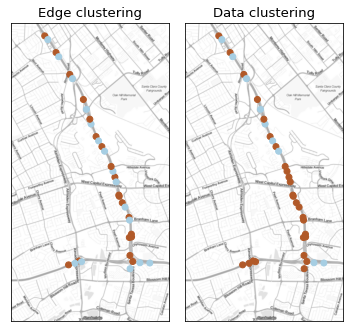} 
    \caption{The figure shows two clusterings of sensors along a freeway in the PEMS dataset. Left: Clustering based on time series learned edges. Right: Clustering based on time series of observed traffic speeds.}
    \label{fig:pems_zoomed_cluster}
\end{figure}

First, we compare the full model against baselines with fixed adjacency matrices,  the DCRNN model from \cite{Li2018DiffusionCR} and a Lag model where the last observed traffic speed is used as a prediction for the next hour. The different models are:
\begin{itemize}
    \setlength\itemsep{0em}
    \item \textbf{DCRNN}: The DCRNN model from \cite{Li2018DiffusionCR}. The values are taken from that paper.
    \item \textbf{Lag}: The last observed traffic speed is repeated for all predictions.
    \item \textbf{Fixed empty}: Model with fixed empty adjacency matrix.
    \item \textbf{Fixed full}: Model with fixed full adjacency matrix.
    \item \textbf{Fixed local}: Model with adjacency matrix based on distance between sensors. The distance between sensors is taken from \cite{Li2018DiffusionCR}.
    \item \textbf{NRI Local prior}: Full NRI model with the local adjacency matrix explained above as prior.
\end{itemize}

The results can be seen in Table \ref{tab:pems}. First, we note that all of our models outperform the DCRNN and Lag models. We also note that the Lag model outperforms the DCRNN model for 15- and 30-minute predictions, but the DCRNN is better for 60-minute predictions\footnote{We also note that the authors of \cite{Li2018DiffusionCR} say that the smooth predictions of the DCRNN model are a feature. However, smooth predictions will lead to lower prediction scores in the selected metrics.}. All of the models with fixed adjacency matrices outperform the Lag and DCRNN models, even the model with an empty adjacency matrix. This shows that the recurrent decoder is already a strong prediction model even without any sharing between nodes. However, the Fixed-local model outperforms the Fixed-empty model showing the benefit of sending information between nodes. Furthermore, the Fixed-full model outperforms the Fixed-local model, showing that the connections in the adjacency matrix based on locality heuristics are not a perfect graph for this problem. The full NRI model achieves similar scores as the Fixed-full model. However, the NRI model has a much more sparse and potentially interpretable adjacency matrix suggesting that the NRI model is able to learn which connections between sensors are important for precise predictions.

\begin{figure}[]
    \centering
    \includegraphics[width=1\columnwidth]{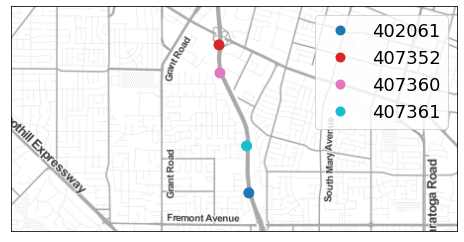} 
    \caption{The mean edge probability for in-going and out-going edges along with the mean traffic speed and traffic speed at the specific sensor for four sensors along the same freeway. As can be seen, a lower than usual traffic speed at a sensor leads to more in-going edges at the sensor.}
    \label{fig:pems_cont_loops}
\end{figure}

Next, we look at the time series of mean edge probabilities to see if we can see a correlation between the data and the graph inferred by the encoder. If we look at Figure \ref{fig:pems_cont_loops}, we can see the observed traffic speed, the mean in-going and out-going edge probability for four loop detectors placed along the same freeway in the southbound direction. The first sensor 407352 is on Freeway 85 just before the merge with traffic from the crossing westbound El Camino Real. The second sensor (407360) is just after the merge, while the third (407361) and fourth (402061) are further south along the freeway. From the plot, we can see that the first and last sensors have regular traffic for the depicted time period. Looking at the mean in-going and out-going probabilities, we can see that they are noisy but maintain a probability of around 30\%. However, looking at the second sensor (407360), we can see that during the Saturday, there was some congestion lowering the traffic speed. At the beginning of the congestion, we can see that that the node first drop and then peak in out-going edge probability. After that, we can see that during the congestion, the mean in-going edge probability for the zone is higher, meaning that the node is taking in more information from the rest of the traffic network when predicting the traffic speed at the sensor. Looking at the third zone, we can see that there is a lesser amount of congestion leading to slightly lowered traffic speeds. This then results in slightly raised in-going edge probabilities. This suggests that the model sends more information to congested zones; the more congested, the more information is sent to that zone. \\

\begin{figure*}[]
    \centering
    \includegraphics[width=1\textwidth]{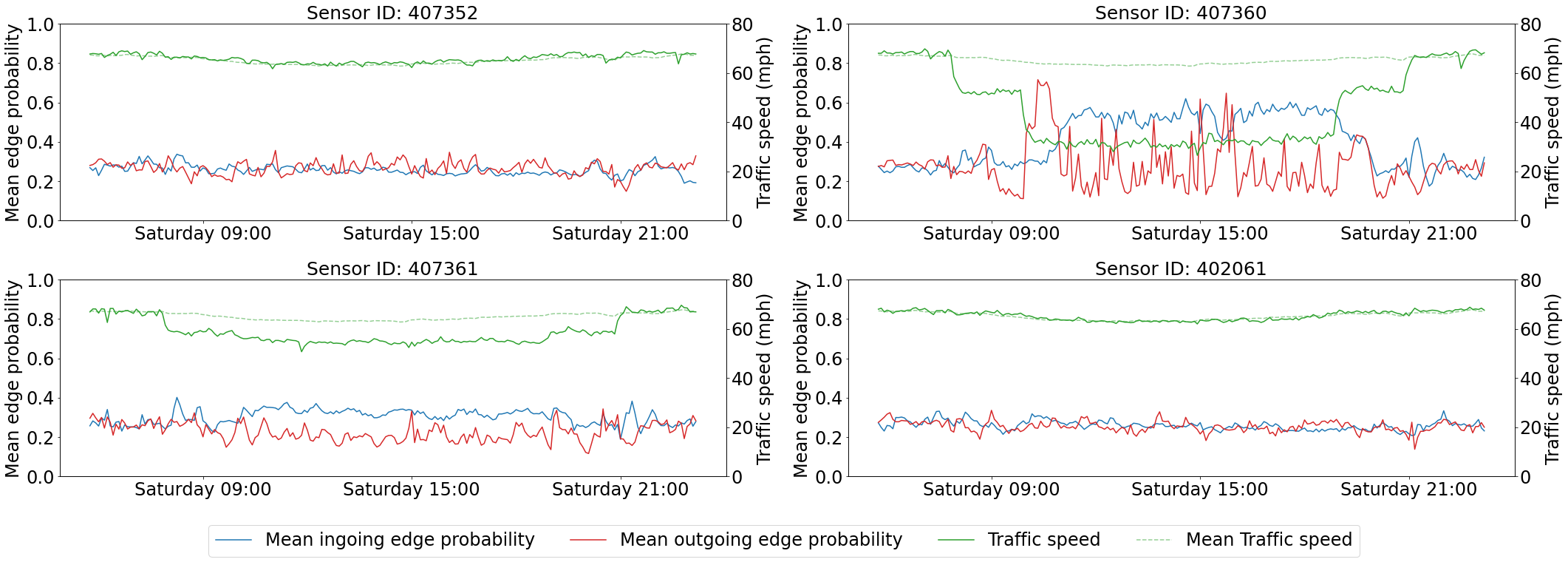} 
    \caption{The mean edge probability for in-going and out-going edges along with the mean traffic speed and traffic speed at the specific sensor for four sensors along the same freeway. As can be seen, a lower than usual traffic speed at a sensor leads to more in-going edges at the sensor.}
    \label{fig:pems_cont_loops_data}
\end{figure*}

In Figure \ref{fig:pems_overall_cluster} and Figure \ref{fig:pems_zoomed_cluster}, we look at clusterings of the loop detectors. From Figure \ref{fig:pems_overall_cluster} we can see that clustering based on the observed and based on the learned connections both leads to clusterings that separate the two directions on the freeway. However, looking at Figure \ref{fig:pems_zoomed_cluster}, we can see that the clustering based on the learned edges leads to better separation of the two directions. This is most clear at the bottom left of the figure.

\begin{figure}[]
    \centering
    \includegraphics[width=0.8\columnwidth]{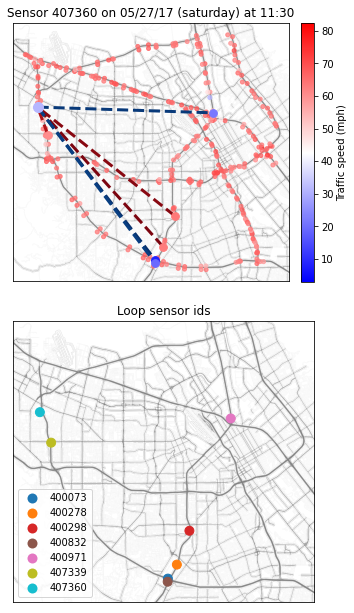} 
    \caption{Top: Learned edges with probability higher than 80\% for sensor 407360 at 11:30 on 05/27/17, which is a Saturday. Blue edge means that the edge is in-going for sensor 407630, and red means out-going. Bottom: Sensor ids for the connected sensors.}
    \label{fig:pems_slow_cons}
\end{figure}

In Figure \ref{fig:pems_slow_cons}, the edges with probability higher than 80\% for sensor 407360 at 11:30 on the Saturday in the test set are depicted. At this time, the freeway around the sensor is congested, and the traffic speed is lowered. As can be seen in the figure, the nodes receive information from three different nodes and send information to 2 other nodes. Interestingly, these nodes are not the immediate neighbors of the node, which could make sense. However, the data has a time fidelity of 5 minutes, and we are predicting the next 12 time steps, i.e., the next hour. Hence the information in the immediate neighbors might be outdated already for the next time step and definitely will be at some point along the prediction window. The nodes the sensor is trying to get information from are all $\sim$12 minutes away\footnote{According to the Google Maps API.}; hence the information at these sensors might be more important for the full prediction window.

\begin{figure}[]
    \centering
    \includegraphics[width=1\columnwidth]{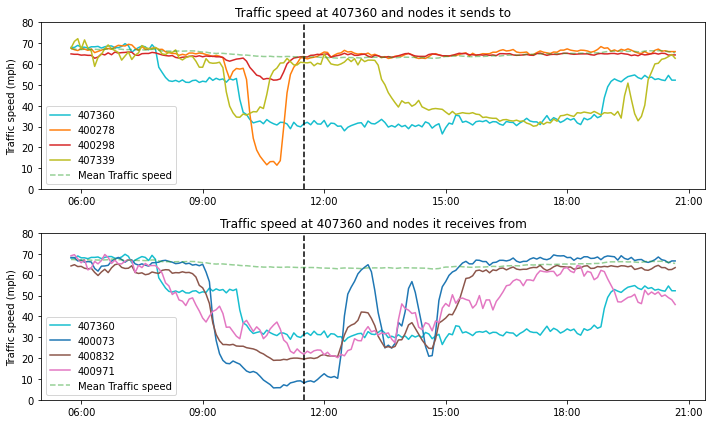} 
    \caption{The traffic speed, mean ingoing and mean outgoing edge probabilities at sensor 407360 and the sensors that 407360 are connected to with an edge with more than 80\% probability. The top figure shows zones connected to via an outgoing edge, and the bottom figure shows zones connected via an ingoing edge.}
    \label{fig:pems_slow_data_simple}
\end{figure}

In the top of Figure \ref{fig:pems_slow_data_simple}, the traffic speed and mean in-going and out-going edge probabilities for sensor 407360 around 11:30 on Saturday is depicted along with the same information for the sensors that 407360 sends information to with a probability higher than 80\%. As can be seen from the figure, the zones that 407360 sends information to are zones that have just had some congestion. In the bottom of Figure \ref{fig:pems_slow_data_simple}, the traffic speed and mean in-going and out-going edge probabilities for sensor 407360 around 11:30 on Saturday is depicted along with the same information for the sensors that 407360 receives information from with a probability higher than 80\%. As can be seen from the figure, all of the zones that 407360 receives information from are zones with more congestion than 407360 itself. 

\begin{figure}[]
    \centering
    \includegraphics[width=0.8\columnwidth]{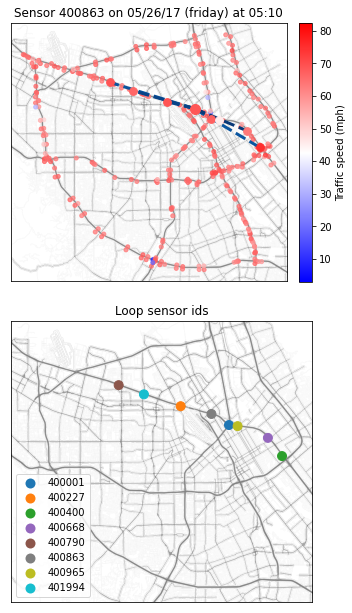} 
    \caption{Top: Learned edges with probability higher than 80\% for sensor 400853 at 5:10 on 05/26/17. Blue edge means that the edge is in-going for sensor 400853. Bottom: Sensor ids for the connected sensors.}
    \label{fig:pems_speed_cons}
\end{figure}

\begin{figure}[]
    \centering
    \includegraphics[width=1\columnwidth]{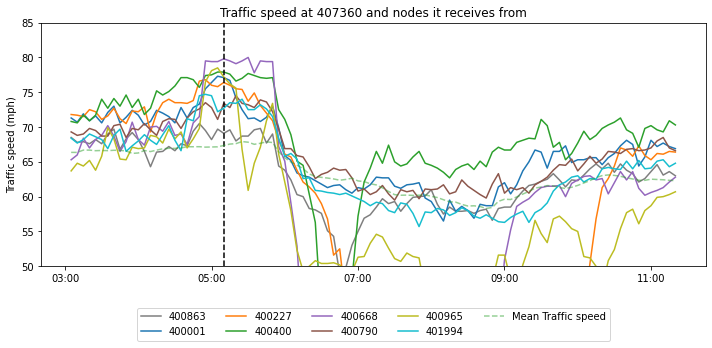} 
    \caption{The traffic speed and mean in-going and out-edge probabilities at senor 400853 and the sensors that 400853 is connected to with a ingoing edge with more than 80\% probability.}
    \label{fig:pems_speed_data}
\end{figure}

In Figure \ref{fig:pems_speed_cons} we depict the sensor 400863 along with the edges with higher than 80\% probability on 05/26/17 at 5:10. As seen in the plot, the sensor only has in-going edges with high probability, and all of these edges are coming from sensors along the same freeway. This shows quite different behavior from Figure \ref{fig:pems_slow_cons}. In Figure \ref{fig:pems_speed_data}, the data at the connected sensors from Figure \ref{fig:pems_speed_cons} is depicted. From the figure it is clear that we are also in a quite different scenario than that around Figure \ref{fig:pems_slow_cons} as all of the connected sensors show higher than average speeds at the time.

Comparing the scenarios in Figure \ref{fig:pems_slow_cons} and \ref{fig:pems_speed_cons}, we observe that the model shows two different connection behaviours. In the scenario with congestion, the model finds other areas with congestion further away to get information from. This makes sense, as congestion further away might contain useful information about the developments at the sensor itself, since congestion spreads along a road network through time. In the free-flow speed scenario, the connections are limited to the same freeway. This also makes sense as speeds exceeding the legal limit along one freeway does not spread to other freeways the same way that congestion does.

\section{Conclusion}
In this paper, we have presented a graph neural network for spatio-temporal demand prediction in transport problems. The model follows the VAE structure from Neural Relational inference and is extended to take in global features. To the authors' knowledge, this is the first model with a learnable adjacency matrix that is not fixed to a specific number of nodes. Furthermore, it is also the first GNN used for transport problems that is able to utilize a proper prior for the learnable adjacency matrix. 

Through experiments on the NYC Yellow Taxicab dataset and the PEMS dataset, we have shown that using a learned, dynamical adjacency matrix can outperform models using fixed adjacency matrices based on common heuristics such as locality and time series correlations. Furthermore, we show that the usage of Bayesian priors on the learned adjacency matrix enables the use of domain knowledge and can increase the model's predictive performance. 

By performing a thorough analysis of the learned adjacency matrices, we show that these matrices are interpretable to some extent and demonstrate how they depend on the learning strategy. For example, we see that in the case of the NYC Yellow Taxicab dataset, the learned adjacency matrix sends the most messages to the zones with the highest amount of activity. This is because these zones have the highest impact on the NLL loss we train the model with. As such, the messages are optimized in aiding predictions in high activity zones, and the adjacency matrix sends these messages to such zones. Another example of the interpretability is the case of the PEMS dataset nodes with congestion. They receive information from nodes with more congestion, even if farther away in space and time. 

In future work, we plan to use the proposed model in a meta learning setup. Since the model is not constrained to a specific number of nodes the model can be used in setups where it is pre-trained using a big data set and then fine tuned and evaluated on a smaller data set. 

\section*{Acknowledgment}
This project has received funding from the European Union’s Horizon 2020 research and innovation programme under grant agreement No 875530.
\bibliography{aaai22.bib}
\end{document}